\documentclass{Interspeech}

\interspeechcameraready

\usepackage[capitalize]{cleveref}

\title{Bridging Audio and Vision: Zero-Shot Audiovisual Segmentation by Connecting Pretrained Models}

\author{Seung-jae}{Lee}
\author{Paul Hongsuck}{Seo}
\affiliation[nocounter]{Department of Computer Science and Engineering}{Korea University}{South Korea}

\email{cid2rrrr@korea.ac.kr, phseo@korea.ac.kr}

\usepackage{comment}
\usepackage{xcolor}
\usepackage{amsmath}
\usepackage{array}
\usepackage{multirow}
\usepackage{arydshln}
\usepackage{booktabs}
\usepackage{graphicx} 
\usepackage{subcaption}
\usepackage{hyperref} 

\begin{document}

\maketitle

\begin{abstract}
Audiovisual segmentation (AVS) aims to identify visual regions corresponding to sound sources, playing a vital role in video understanding, surveillance, and human-computer interaction. Traditional AVS methods depend on large-scale pixel-level annotations, which are costly and time-consuming to obtain. To address this, we propose a novel zero-shot AVS framework that eliminates task-specific training by leveraging multiple pretrained models. Our approach integrates audio, vision, and text representations to bridge modality gaps, enabling precise sound source segmentation without AVS-specific annotations. We systematically explore different strategies for connecting pretrained models and evaluate their efficacy across multiple datasets. Experimental results demonstrate that our framework achieves state-of-the-art zero-shot AVS performance, highlighting the effectiveness of multimodal model integration for fine-grained audiovisual segmentation.{\footnote{The codes are avaliable at \url{https://cid2rrrr.github.io/BAaV/}}}
\end{abstract}

\section{Introduction}

Audiovisual segmentation (AVS) is a fundamental task in multimodal learning, that aims to localize visual regions that correspond to sound sources through the alignment and fusion of audio and visual cues. This capability is essential for applications in video understanding, surveillance, human-computer interaction, and autonomous systems. By effectively segmenting sound-emitting objects, AVS enables machines to better understand complex audiovisual scenes, enhancing both perception and interaction. Traditional AVS approaches rely on large-scale datasets with pixel-level annotations to achieve high performance. However, acquiring such annotations is labor-intensive, costly, and often impractical at scale, which limits the applicability of supervised approaches in real-world settings.

To address these challenges, we propose a novel zero-shot AVS framework that eliminates the need for task-specific training by leveraging pretrained models. Instead of relying on AVS-specific annotations, our approach systematically integrates multiple pretrained models from different subdomains, including audio, vision, and text, to enable cross-modal reasoning without additional annotations. By harnessing the complementary strengths of different pretrained models, our framework achieves accurate audiovisual segmentation in a zero-shot setting.

Specifically, we investigate multiple strategies for integrating pretrained models to enable zero-shot AVS. In particular, we examine different approaches to bridge audio and visual modalities through text representations, leveraging recent advances in multimodal learning. Our experiments demonstrate that our framework significantly outperforms existing unsupervised AVS methods, achieving state-of-the-art zero-shot performance. These findings suggest that effective multimodal reasoning—achieved by systematic composition of pretrained models—can open new directions for learning-based audiovisual scene understanding.

Our contributions are threefold: (1) We propose a novel approach to zero-shot AVS that leverages pretrained models from related tasks. (2) We systematically explore multiple strategies for integrating these pretrained models, conducting extensive evaluations. (3) Our final model achieves state-of-the-art zero-shot AVS performance, demonstrating the effectiveness of multimodal model fusion in this task.

\section{Related Works}

Sound Source Localization (SSL) is a fundamental task in audiovisual learning that aims to localize sound-emitting objects within visual scenes. Many approaches leverage cross-modal attention~\cite{llsvs,sslalign,ssltie} combined with contrastive learning~\cite{ezvsl,lvs} to align audio and visual features. To improve robustness, various strategies such as noise contrastive estimation~\cite{fnac}, hard negative mining~\cite{lvs}, and transformation-invariant training~\cite{ssltie} have been proposed. Additional enhancements include negative-margin contrastive learning~\cite{marginnce} and the use of multiple positive pairs~\cite{sslalign} for better cross-modal alignment. However, SSL often results in coarse localization without fine object boundaries.

To address this limitation, Audio-Visual Segmentation (AVS) focuses on generating pixel-level masks of sounding objects. The introduction of AVSBench~\cite{avsbench} enabled rigorous benchmarking, leading to methods such as bidirectional reconstruction~\cite{avsbid}, selective suppression~\cite{selm}, and the use of foundation models~\cite{bavs}. Despite these advances, existing AVS models remain reliant on task-specific training. In contrast, our work explores a training-free zero-shot framework that integrates pretrained audio and vision models via textual representations, highlighting the potential of open-domain priors for scalable and robust multimodal segmentation.

\section{Proposed Method}

\subsection{Audiovisual Segmentation}

Audiovisual Segmentation (AVS) aims to identify and segment regions in a visual frame that correspond to sound-emitting objects, by jointly analyzing audio and visual modalities.
Given a visual frame with synchronized audio, the goal is to generate a binary segmentation mask highlighting the spatial regions in the image corresponding to the sound source.
Formally, let $I \in \mathbb{R}^{H\times W \times 3}$ be a visual frame, with height $H$ and width $W$, and let $A \in \mathbb{R}^{L}$ be an audio synchronized with $I$, where $L$ denotes the number of audio samples.
Given an audiovisual input pair $(I,A)$, a model predicts a binary segmentation mask $M \in \{0,1\}^{H \times W}$, where $M(i,j)=1$ indicates that the pixel at location $(i,j)$ belongs to a sound-emitting region, and $M(i,j)=0$ otherwise.

\subsection{Zero-Shot Audiovisual Segmentation Approaches} \label{subsec:approach}
A key challenge in AVS is modeling the interactions between auditory and visual modalities. 
While supervised approaches rely on large-scale datasets of audiovisual streams annotated with dense segmentation maps, such data are scarce and expensive to obtain.
In contrast, extensive research~\cite{clip,msclap} and abundant data~\cite{wavcaps,audiocaps,mscoco} exist for audio-text and vision-text associations, resulting in strong pretrained models in these domains. 
Building on these advancements, we propose zero-shot AVS strategies that interconnect multiple pretrained models, enabling segmentation without task-specific annotations. 
By leveraging existing multimodal models, our approach circumvents the need for labor-intensive data collection while achieving fine-grained segmentation in AVS.

Notably, many existing pretrained models operate within a text-anchored representation space—for example, CLAP~\cite{msclap} and WavCaps~\cite{wavcaps} align audio with text, while BLIP~\cite{blip} connects visual inputs with natural language. This observation allows us to utilize the text modality as a shared intermediary for bridging different models.
Specifically, in this work, we explore and systematically compare multiple potential approaches for integrating various pretrained models to achieve zero-shot AVS.
A core component across all approaches is a pretrained model for referring image segmentation (RIS), which identifies the region mask $M$ corresponding to a referring textual description $d$ within the image $I$.
This is essential, as our target task aims to generate a fine-grained segmentation map.
Given this framework, the primary design challenge becomes constructing an appropriate textual representation $d$ from the audiovisual input pair $(I, A)$—a process that differentiates the approaches detailed below.

\noindent\textbf{Audio Classification} \ \ 
Since our goal is to identify the visual region of the sound source, a na\"ive approach would be to classify the audio signals $A$ to obtain a textual label (\textit{e.g.}, `vehicle' or `dog') representing the sound source.
This label can then be used to form the input $d$ for the RIS model.
Specifically, we use the common prompt `a photo of $\{c\}$.' where $c$ is the predicted class label.

\noindent\textbf{Audio Captioning} \ \ 
An alternative approach to bridging the audio and text modalities is to generate a \textit{natural language description} rather than a predefined class label. For example, a caption such as “a girl is singing” provides richer contextual and semantic information compared to a single-word label like “girl,” potentially offering improved grounding for identifying the relevant visual region.

\noindent\textbf{Text Inversion} \ \ 
While the former two approaches convert audio into explicit textual form, one can alternatively utilize a feature representation to bridge the audio and text modalities.
Specifically, we can leverage an audio-text alignment model such as CLAP~\cite{msclap}, which embeds audio and text features within a shared space in a semantically aligned manner.
By training such a model with a frozen RIS text encoder, we can estimate text token embeddings $\hat{\mathbf{e}}_d$ of $d$, which is unknown, that correspond to the input audio $A$ through textual inversion~\cite{inversion}.
Formally, we estimate $\hat{\mathbf{e}}_d$ by 
\begin{equation}
    \hat{\mathbf{e}}_d = \underset{\mathbf{e}_d}{\mathrm{argmin}} \, \phi(\Psi_a(A), \Psi_t(\mathbf{e}_d))
\end{equation}
where $\Psi_a$ is the audio encoder, $\Psi_t$ is the text encoder without the embedding layer, $\phi(a, b)$ denotes the cosine similarity of $a$ and $b$, and $\hat{\mathbf{e}}_d$ is the estimated text embeddings.
Once $\hat{\mathbf{e}}_d$ is obtained, it is directly injected into the RIS model as the text representation, bypassing the standard text embedding stage.

\begin{figure}[t]
    \centering
    \begin{subfigure}{\linewidth}
        \centering
        \includegraphics[width=\linewidth]{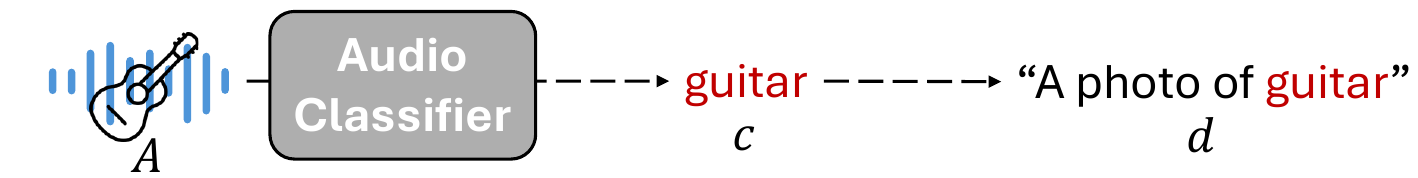}%
        \vspace{-0.1cm}
        \caption{Audio Classification}
        \vspace{0.4cm}
        \label{fig:subfig1}%
    \end{subfigure}
    
    \begin{subfigure}{\linewidth}
        \centering
        \includegraphics[width=\linewidth]{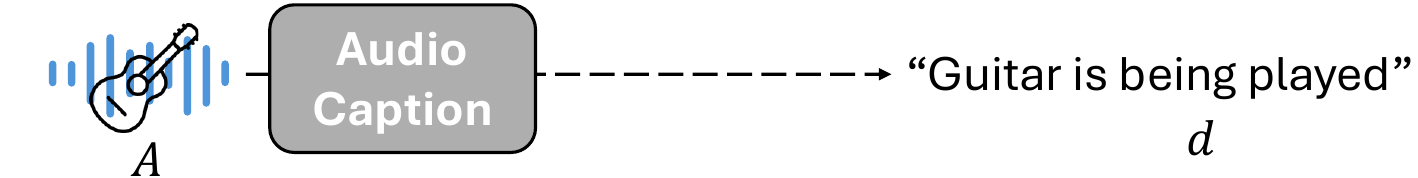}%
        \vspace{-0.1cm}
        \caption{Audio Captioning}
        \vspace{0.4cm}
        \label{fig:subfig2}%
    \end{subfigure}
    
    \begin{subfigure}{\linewidth}
        \centering
        \includegraphics[width=\linewidth]{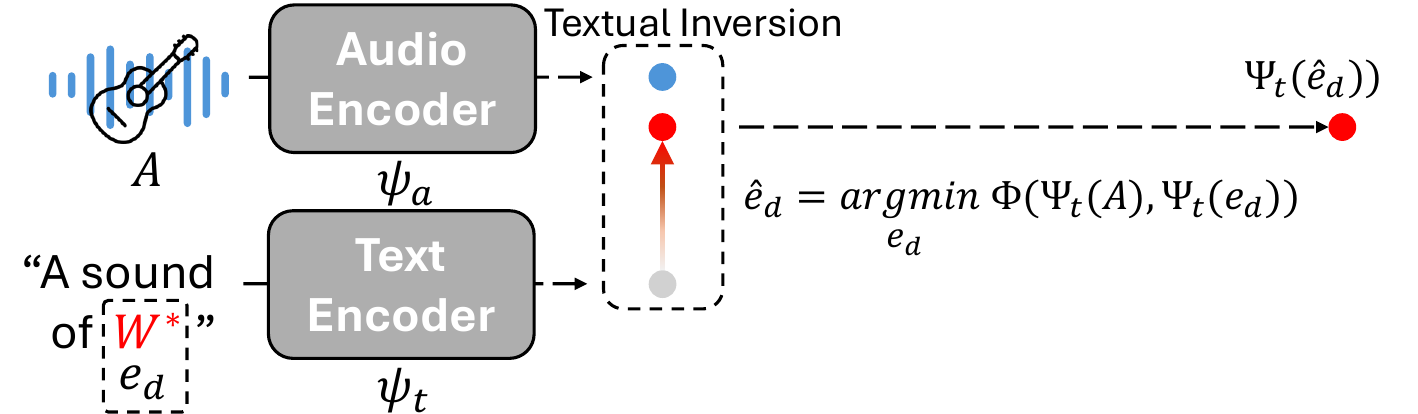}%
        \vspace{-0.1cm}
        \caption{Text Inversion}
        \vspace{0.4cm}
        \label{fig:subfig3}%
    \end{subfigure}
    
    \begin{subfigure}{\linewidth}
        \centering
        \includegraphics[width=\linewidth]{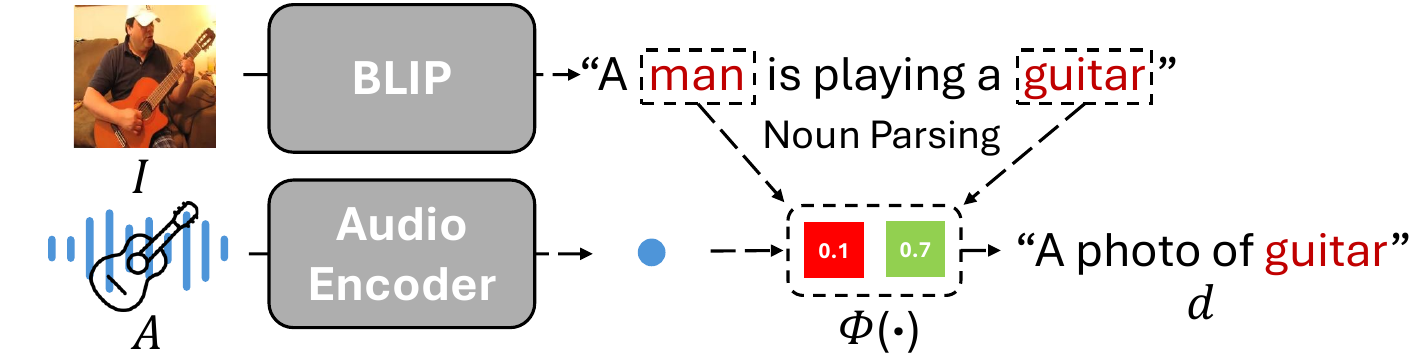}%
        \caption{Cross-modal Verification}
        \label{fig:subfig4}%
    \end{subfigure}
    
    
    
    \caption{Overview of proposed zero-shot AVS approaches. Each subfigure shows a strategy to convert audiovisual inputs into textual queries for the RIS model (see~\cref{subsec:approach}).}
    \label{fig:frameworks} 
\end{figure}

\noindent\textbf{Cross-modal Verification} \ \ 
While AVS is a multimodal task that connects audio and vision, the above approaches primarily focus on audio-only recognition, leaving visual understanding entirely to the RIS model. 
However, visual context can provide crucial context for interpreting audio signals, particularly when distinguishing between similar sounds (\textit{e.g.}, distinguishing a buzzing bee from an electric shaver)~\cite{learning}.
To leverage this insight, we propose a cross-modal verification mechanism that jointly incorporate both the image $I$ and audio $A$.
Specifically, we employ an image captioning model to identify visual objects $I$, which helps constrain the audio classification model by rejecting scores for objects that are not visually present. 
This multimodal verification step effectively reduces candidate sound classes to only those that align with the visual scene, eliminating ambiguous cases that can be clearly ruled out based on visual evidence (\textit{e.g.}, rejecting the bee class when no bee is visible).
Alternatively, this process can be reversed by incorporating audio captioning and image classification models, where the identified audio candidates help filter out inconsistent classes in the image classification step.

\section{Experiments}

\subsection{Models}
Below, we elaborate on the final models built for each approach. 
For all models, we use the same RIS model, ASDA~\cite{asda}.

\noindent \textbf{Audio Classification} \ \ 
For this approach, we use BEATs~\cite{Chen2022beats}, a classifier pretrained on AudioSet~\cite{audioset}. 

\noindent \textbf{Audio Captioning} \ \ 
We use WavCap~\cite{wavcaps} pretrained on the Audiocaps~\cite{audiocaps}, Clotho~\cite{clotho}, and WavCaps dataset~\cite{wavcaps}.

\noindent \textbf{Text Inversion} \ \ 
We employ CLAP~\cite{msclap}, trained with a frozen CLIP~\cite{clip} text encoder, ensuring that the model shares the same text encoder as the RIS model.

\noindent \textbf{Cross-modal Verification} \ \ 
We use BLIP~\cite{blip} for image captioning and CLAP~\cite{msclap} for open-vocabulary classification.
Specifically, after BLIP generates a caption, we extract noun phrases using SpaCy~\cite{spacy} and use them as class labels for CLAP's audio classification.
In addition, we employ WavCaps~\cite{wavcaps} and CLIP~\cite{clip} for the reverse process.

\subsection{Datasets}
We validate our proposed approaches on the following three datasets for AVS. 
Note that our method is entirely zero-shot and thus is evaluated exclusively on the test sets.

\noindent \textbf{IS3} \ \ 
IS3~\cite{sslalign} is a large-scale test dataset for AVS with synthesized images. 
Each image is generated using Stable Diffusion~\cite{stablediffusion}, with prompts derived from the class labels of two randomly selected audio samples from VGGSound~\cite{vggsound}. 
The GT (Ground Truth) segmentation mask for each image-audio pair is manually selected from the mask proposals generated by SAM~\cite{sam}.
The dataset comprises 3,240 examples, each containing an image paired with two audio samples, each corresponding to a different sound source with its own GT mask.

\noindent \textbf{AVSBench-S4} \ \ 
We also evaluate our approaches on AVSBench-S4, a subset of AVSBench~\cite{avsbench} containing video samples with a single sound source.
It consists of manually curated 5-second videos with synchronized audio sourced from YouTube and provides pixel-level dense segmentation masks at 1 fps.
The dataset contains 740 videos in the test split.

\noindent \textbf{VPO-SS} \ \ 
VPO-SS is a subset of the VPO dataset~\cite{vpo} containing a single sound source. 
Each example is automatically constructed by pairing an image and segmentation mask from MS-COCO~\cite{mscoco} with an audio sample from VGGSound~\cite{vggsound} based on their annotated labels. 
The test split includes 890 samples.

\subsection{Metrics}
For evaluation, we adopt mIoU, cIoU, AUC, and F-score, as introduced in~\cite{llsvs}. While these metrics were originally measured using the top 50\% of pixels with the highest segmentation scores, we follow the adaptive pixel selection approach from~\cite{sslalign}. 
This method instead selects the top-$k$ pixels, where $k$ corresponds to the number of pixels in the GT mask.

\noindent \textbf{mIoU} \ \ 
mIoU computes the average Intersection over Union (IoU) between predicted and GT masks across all samples. 

\noindent \textbf{cIoU} \ \ 
This metric measures the proportion of samples that achieve an IoU score greater than 0.5.

\noindent \textbf{AUC} \ \ 
AUC is computed by varying the IoU threshold used in cIoU and measuring the area under the resulting curve.

\noindent \textbf{F-score} \ \ 
F-score balances precision and recall  with $\beta=0.3$ comparing predicted and GT masks at pixel-level.

\noindent \textbf{$\mathcal{J}$ and $\mathcal{F}$} \ \ 
In addition to the above metrics that rely on GT mask sizes, we evaluate the mIoU ($\mathcal{J}$) and F-score ($\mathcal{F}$) of the final predicted masks using thresholding~\cite{asda}.
For our model, we adopt the thresholds selected for the original RIS models.

\begin{table}[t]
\centering
\caption{Performance Comparison of proposed zero-shot AVS strategies across multiple datasets. Our VCap+ACls (Cross-modal Verification) approach consistently achieves the highest performance across all metrics. See~\cref{subsec:result} for details.}
\scalebox{0.85}{
\begin{tabular}{c|l|cccccc}
\toprule
 & Model & cIoU & AUC & mIoU & Fscore & $\mathcal{J}$ & $\mathcal{F}$ \\
\midrule
\multirow{5}{*}{\rotatebox{90}{IS3}} 
 & Classification      & 58.3 & 53.8 & 53.2 & 59.3 & 48.3 & 54.4 \\
 & Captioning    & 56.3 & 52.0 & 51.5 & 58.1 & 44.1 & 51.3 \\
 & Inversion      & 35.5 & 35.0 & 34.2 & 40.9 & 23.3 & 29.7 \\
 & {VCap+ACls} & \textbf{65.4} & \textbf{59.6} & \textbf{59.1} & \textbf{65.2} & \textbf{54.2} & \textbf{60.7} \\
 & ACap+VCls             & 62.8 & 57.8 & 57.3 & 63.5 & 52.5 & 58.8 \\
\midrule
\multirow{5}{*}{\rotatebox{90}{AVSBench-S4}} 
 & Classification      & 59.8 & 53.5 & 53.0 & 59.6 & 48.0 & 56.1 \\
 & Captioning    & 63.8 & 56.7 & 56.2 & 63.1 & 48.9 & 58.4 \\
 & Inversion      & 31.8 & 32.8 & 32.2 & 39.3 & 19.2 & 26.0 \\
 & {VCap+ACls} & \textbf{71.1} & \textbf{62.4} & \textbf{62.0} & \textbf{69.2} & \textbf{57.0} & \textbf{66.0} \\
 & ACap+VCls             & 70.2 & 61.5 & 61.1 & 68.1 & 55.6 & 64.7 \\
\midrule
\multirow{5}{*}{\rotatebox{90}{VPO-SS}} 
 & Classification      & 42.2 & 35.7 & 34.9 & 43.4 & 27.7 & 38.5 \\
 & Captioning    & 48.4 & 39.3 & 38.6 & 47.6 & 33.2 & 45.7 \\
 & Inversion      & 24.2 & 25.0 & 24.1 & 31.9 & 13.8 & 21.2 \\
 & {VCap+ACls} & \textbf{54.6} & \textbf{49.3} & \textbf{48.5} & \textbf{52.7} & \textbf{46.0} & \textbf{50.1} \\
 & ACap+VCls             & 49.8 & 39.4 & 38.6 & 47.2 & 34.9 & 47.4 \\
\bottomrule
\end{tabular}
}
\label{tab:ours_compare}
\end{table}

\subsection{Results} \label{subsec:result}

\noindent \textbf{Comparisons of Proposed Zero-Shot Approaches} \ \ 
In \cref{tab:ours_compare}, we compare five zero-shot AVS strategies we propose.
Despite their simplicity, both Classification and Captioning achieve strong performances across all three datasets.
Interestingly, while Captioning generally outperforms Classification (on AVSBench-S4 and VPO-SS) due to its rich descriptions, it can sometimes confuse the RIS model, resulting in scattered segmentation masks and limited performance in certain domains, such as IS3.
Inversion yields the lowest performance among the proposed methods, likely due to the inherent difficulty of reconstructing well-aligned textual embeddings solely from audio features. 
A key limitation is that
the text encoder—frozen to match the RIS model—was pretrained for vision-language alignment, not audio-text correspondence. Additionally, modality gaps in models like CLIP and CLAP further degrade alignment quality, making it difficult for Inversion to produce meaningful embeddings.

Compared to the approaches that rely solely on the audio modality, Both cross-modal approaches, VCap+ACls and ACap+VCls, that leverage both audio and visual information achieve significantly stronger performance in most configurations.
For instance, VCap+ACls outperforms the best-performing audio-only approach, Classification, by over 5.8\% relative improvements on IS3 across all metrics.
When comparing the two cross-modal methods, VCap+ACls demonstrated significantly higher performances than ACap+VCls.
This is partly because visual captions provide a more reliable and concrete representation of the scene, effectively constraining the audio classification process. 
In contrast, audio captions can be abstract or ambiguous, making it more challenging to filter out inconsistent visual classes.
Note that aggressive rejection of object classes at an early stage can lead to severe error propagation, resulting in significant performance drops.

\begin{table}[t]
\centering
\caption{Performance comparison with state-of-the-art unsupervised AVS methods in a zero-shot setting. Our approach outperforms them without additional training.}
\scalebox{0.85}{
\begin{tabular}{c|l|cccccc}
\toprule
 & Model & cIoU & AUC & mIoU & Fscore & $\mathcal{J}$ & $\mathcal{F}$ \\
\midrule
\multirow{8}{*}{\rotatebox{90}{IS3}} 
& LVS~\cite{lvs}        & 11.1 & 24.2 & 23.9 & 34.8 & 21.2 & 25.7 \\
& EZ-VSL~\cite{ezvsl}     & 13.4 & 26.4 & 26.0 & 37.3 & 20.7 & 25.1 \\
& SSL-TIE~\cite{ssltie}    & 20.7 & 31.8 & 31.5 & 44.1 & 27.8 & 34.6 \\
& SLAVC~\cite{slavc}      & 15.1 & 26.2 & 25.7 & 36.6 & 18.1 & 21.9 \\
& MarginNCE~\cite{marginnce}  & 18.5 & 30.8 & 30.5 & 42.9 & 21.6 & 26.0 \\
& FNAC~\cite{fnac}       & 14.7 & 27.5 & 27.1 & 38.7 & 19.4 & 23.4 \\
& SSLalign~\cite{sslalign}   & 28.5 & 36.6 & 36.4 & 49.8 & 31.0 & 38.1 \\
\cdashline{2-8}
& \textbf{Ours}  & \textbf{65.4} & \textbf{59.6} & \textbf{59.1} & \textbf{65.2} & \textbf{54.2} & \textbf{60.7} \\
\midrule
\multirow{8}{*}{\rotatebox{90}{AVSBench-S4}} 
& LVS~\cite{lvs}        & 21.1 & 30.6 & 30.5 & 42.4 & 24.4 & 29.2 \\
& EZ-VSL~\cite{ezvsl}     & 19.5 & 30.8 & 30.7 & 42.8 & 23.0 & 27.3 \\
& SSL-TIE~\cite{ssltie}    & 31.1 & 38.5 & 28.9 & 52.5 & 20.3 & 26.2 \\
& SLAVC~\cite{slavc}      & 21.8 & 32.8 & 32.8 & 45.5 & 21.1 & 25.1 \\
& MarginNCE~\cite{marginnce}  & 25.2 & 35.5 & 35.4 & 48.6 & 23.9 & 28.4 \\
& FNAC~\cite{fnac}       & 23.0 & 33.1 & 33.0 & 45.6 & 22.5 & 26.7 \\
& SSLalign~\cite{sslalign}   & 30.6 & 39.2 & 39.2 & 53.0 & 32.2 & 38.8 \\
\cdashline{2-8}
& \textbf{Ours}  & \textbf{71.1} & \textbf{62.4} & \textbf{62.0} & \textbf{69.2} & \textbf{57.0} & \textbf{66.0} \\
\midrule
\multirow{8}{*}{\rotatebox{90}{VPO-SS}} 
& LVS~\cite{lvs}        & 14.6 & 21.4 & 20.5 & 28.6 & 18.8 & 22.4 \\
& EZ-VSL~\cite{ezvsl}     & 12.2 & 22.1 & 21.2 & 29.9 & 18.3 & 21.9 \\
& SSL-TIE~\cite{ssltie}    & 20.4 & 26.3 & 25.4 & 34.8 & 21.3 & 27.7 \\
& SLAVC~\cite{slavc}      & 15.3 & 22.1 & 21.2 & 29.5 & 17.1 & 20.2 \\
& MarginNCE~\cite{marginnce}  & 14.8 & 23.4 & 22.5 & 31.4 & 18.8 & 22.3 \\
& FNAC~\cite{fnac}       & 15.7 & 23.2 & 22.2 & 30.8 & 18.0 & 21.3 \\
& SSLalign~\cite{sslalign}   & 17.6 & 24.9 & 24.1 & 33.3 & 22.0 & 28.0 \\
\cdashline{2-8}
& \textbf{Ours}  & \textbf{54.6} & \textbf{49.3} & \textbf{48.5} & \textbf{52.7} & \textbf{46.0} & \textbf{50.1} \\
\bottomrule
\end{tabular}
}
\label{tab:performance}
\end{table}

\noindent \textbf{Comparisons to SOTA Methods} \ \ 
In this section, we compare our zero-shot AVS approaches with existing unsupervised methods in AVS under a zero-shot setting, including SSL-TIE~\cite{ssltie}, EZ-VSL~\cite{ezvsl}, SLAVC~\cite{slavc}, MarginNCE~\cite{marginnce}, FNAC~\cite{fnac}, and SSLalign~\cite{sslalign}. \cref{tab:performance} presents quantitative comparisons across three benchmark datasets: IS3, AVSBench-S4, and VPO-SS.

Notably, prior methods rely on self-supervised training to learn cross-modal associations, often using contrastive objectives or transformation constraints. However, they tend to produce coarse masks due to low-resolution feature maps and lack of strong semantic grounding. whereas our zero-shot approach circumventsthis by integrating multiple pretrained models from different subdomains, enabling fine-grained segmentation without task-specific training.

Across all datasets, our Cross-modal Verification approach consistently achieves state-of-the-art performance, surpassing existing unsupervised methods across all evaluation metrics by a significant margin. Notably, our method improves performance by approximately 2× across most metrics and attains a 3× improvement in cIoU on the VPO-SS dataset. These results underscore the effectiveness of our Cross-modal Verification approach in harnessing the generalization capabilities of pretrained models without the need for additional training. In contrast, prior methods struggle to generate accurate segmentation masks due to their reliance on low-resolution feature maps and the inherent limitations of self-supervised learning.

\begin{figure}[t]
    \centering
    \includegraphics[width=\linewidth]{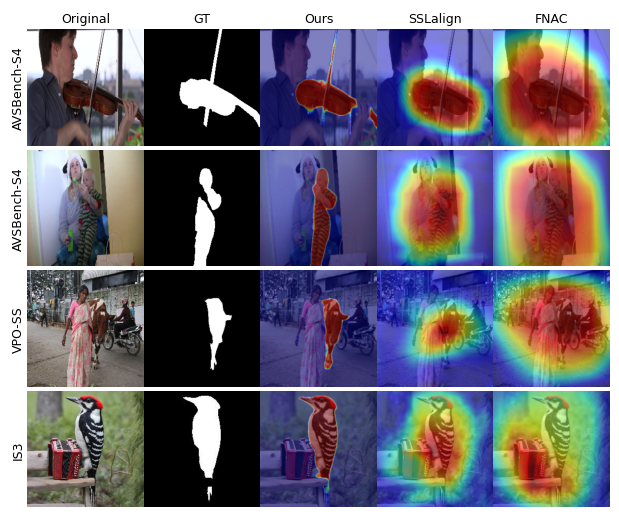}
    \caption{Qualitative comparisons of segmentation results across datasets. Our method produces fine-grained masks for sounding objects, outperforming prior methods in visual accuracy and boundary alignment.}
    \label{fig:example}
\end{figure}

A qualitative comparison, presented in \cref{fig:example}, further highlights the superiority of our method. As previously discussed, existing methods produce overly coarse segmentation masks that frequently cover irrelevant objects within the scene. While these methods attempt to localize sound-emitting regions, their highest activation scores are often misplaced, indicating their difficulty to achieve precise localization.

Our approach effectively utilizes the reasoning capabilities of RIS models through audiovisual Cross-modal Verification. By explicitly decomposing the segmentation process into a framework that leverages visual information to interpret audio cues, our method achieves precise, fine-grained pixel-level segmentation even in complex, multi-object scenarios.

\ \

\begin{table}[t]
\centering
\caption{Ablation study of different RIS models. Our approach with ASDA~\cite{asda} achieves the best performance on all datasets.}
\scalebox{0.85}{
\begin{tabular}{c|l|cccccc}
\toprule
 & Model & cIoU & AUC & mIoU & Fscore & $\mathcal{J}$ & $\mathcal{F}$ \\
\midrule
\multirow{3}{*}{\rotatebox{90}{IS3}} 
 & \textbf{ASDA~\cite{asda}}     & \textbf{65.4} & \textbf{59.6} & \textbf{59.1} & \textbf{65.2} & \textbf{54.2} & \textbf{60.7} \\
 & ETRIS~\cite{etris}    & 64.3 & 57.2 & 56.8 & 64.3 & 52.3 & 60.8 \\
 & CRIS~\cite{cris}     & 53.4 & 47.0 & 46.4 & 53.4 & 42.9 & 50.9 \\
\midrule
\multirow{3}{*}{\rotatebox{90}{S4}}
 & \textbf{ASDA~\cite{asda}}     & \textbf{71.1} & \textbf{62.4} & \textbf{62.0} & \textbf{69.2} & \textbf{57.0} & \textbf{66.0} \\
 & ETRIS~\cite{etris}    & 59.2 & 52.3 & 51.8 & 60.7 & 46.5 & 58.1 \\
 & CRIS~\cite{cris}     & 56.1 & 50.0 & 49.5 & 57.4 & 45.2 & 55.3 \\
\midrule
\multirow{3}{*}{\rotatebox{90}{VPOSS}} 
 & \textbf{ASDA~\cite{asda}}     & \textbf{65.4} & \textbf{59.6} & \textbf{59.1} & \textbf{65.2} & \textbf{54.2} & \textbf{60.7} \\
 & ETRIS~\cite{etris}    & 53.7 & 46.4 & 45.6 & 50.7 & 43.2 & 48.9 \\
 & CRIS~\cite{cris}     & 51.8 & 45.9 & 45.0 & 49.9 & 42.9 & 48.6 \\
\bottomrule
\end{tabular}
}
\label{tab:dummy}
\end{table}

\noindent \textbf{Ablations with Various RIS Models} \ \ 
To evaluate the impact of RIS models on AVS performance, we conduct an ablation study where we replace the RIS model with different architectures. \cref{tab:dummy} compares our default RIS model, ASDA~\cite{asda}—a state-of-the-art method for RIS—with two alternatives: ETRIS~\cite{etris} and CRIS~\cite{cris}. Our findings indicate that ASDA consistently outperforms across all datasets. This suggests that the choice of RIS model significantly influences segmentation performance. These results emphasize that selecting a high-quality RIS model is essential for achieving optimal zero-shot AVS performance.

\section{Conclusion}

Our study introduces a novel zero-shot audiovisual segmentation framework that obviates the need for task-specific training by leveraging multiple pretrained models. By systematically integrating audio, vision, and text representations, we bridge modality gaps and achieve fine-grained segmentation without requiring AVS-specific annotations. Our approach outperforms existing unsupervised methods, demonstrating the effectiveness of multimodal model fusion for audiovisual scene understanding. Furthermore, our findings highlight the critical role of pretrained model selection, particularly in RIS, in ensuring high segmentation accuracy. This work opens new directions for zero-shot audiovisual segmentation, paving the way for scalable and efficient multimodal learning in real-world applications.

\vfill
\clearpage

\section{Acknowledgements}
This research was supported by IITP grants
(IITP\allowbreak-2025-RS-2020-II201819, 
IITP-2025-RS-2024-00436857, 
IITP-2025-RS-2024-00398115, 
IITP-2025-RS-2025-02263754, 
IITP-2025-RS-\allowbreak2025-02304828
) and the KOCCA grant (RS-2024-00345025
) funded by the Korea government (MSIT, MOE and MSCT).

\bibliographystyle{IEEEtran}
\bibliography{ref}

\end{document}